# DCAN: Dual Channel-wise Alignment Networks for Unsupervised Scene Adaptation

Zuxuan Wu[1], Xintong Han[1], Yen-Liang Lin[2], Mustafa Gkhan Uzunbas[2], Tom Goldstein[1], Ser Nam Lim[2], Larry S. Davis[1]

[1]University of Maryland, College Park    [2]GE Global Research

**Abstract.** Harvesting dense pixel-level annotations to train deep neural networks for semantic segmentation is extremely expensive and unwieldy at scale. While learning from synthetic data where labels are readily available sounds promising, performance degrades significantly when testing on novel realistic data due to domain discrepancies. We present Dual Channel-wise Alignment Networks (DCAN), a simple yet effective approach to reduce domain shift at both pixel-level and feature-level. Exploring statistics in each channel of CNN feature maps, our framework performs channel-wise feature alignment, which preserves spatial structures and semantic information, in both an image generator and a segmentation network. In particular, given an image from the source domain and unlabeled samples from the target domain, the generator synthesizes new images on-the-fly to resemble samples from the target domain in appearance and the segmentation network further refines high-level features before predicting semantic maps, both of which leverage feature statistics of sampled images from the target domain. Unlike much recent and concurrent work relying on adversarial training, our framework is lightweight and easy to train. Extensive experiments on adapting models trained on synthetic segmentation benchmarks to real urban scenes demonstrate the effectiveness of the proposed framework.

## 1 Introduction

Deep neural networks have driven recent advances in computer vision. However, significant boosts in accuracy achieved by high-capacity deep models require large corpora of manually labeled data such as ImageNet [1] and COCO [2]. The need to harvest clean and massive annotations limits the ability of these approaches to scale, especially for fine-grained understanding tasks like semantic segmentation, where dense annotations are extremely costly and time-consuming to obtain. One possible solution is to learn from synthetic images rendered by modern computer graphics tools (*e.g.*, video game engines), such that ground-truth labels are readily available. While synthetic data have been exploited to train deep networks for a multitude of tasks like depth estimation [3], object detection [4], *etc.*, the resulting models usually suffer from poor generalization when exposed to novel realistic samples. The reasons are mainly two-folds: (1) the realism of synthesized images is limited—inducing an inherent gap between synthetic and real image distributions; (2) deep networks are prone to overfitting in the training stage, which leads to limited generalization ability.

Learning a discriminative model that reduces the disparity between training and testing distributions is typically known as domain adaptation; a more challenging setting is unsupervised domain adaptation that aims to bridge the gap without accessing labels of the testing domain during training. Most existing work seeks to align features in a deep network of the *source* domain (training sets) and the *target* domain (testing sets) by either explicitly matching feature statistics [5,6,7] or implicitly making features domain invariant [8,9]. Recent work also attempts to minimize domain shift in the pixel space to make raw images look alike [10,11,12] with adversarial training. While good progress has been made for classification, generalizing these ideas to semantic segmentation has been shown to be less effective [13], possibly due to the fact that high-dimensional feature maps are more challenging to align compared to features used for classification from fully-connected layers.

In this paper, we study unsupervised domain adaptation for semantic segmentation, which we refer as unsupervised scene adaptation. We posit that channel-wise alignment of high-level feature maps is important for adapting segmentation models, as it is able to preserve spatial structures and consider semantic information like attributes and concepts encoded in different channels [14] independently, which implicitly helps transfer feature distributions between the corresponding concepts across domains. In particular, we build upon recent advances of instance normalization [15] due to its effectiveness and simplicity for style transfer [15,16,17]. Instance normalization is motivated by the fact that mean and standard deviation in each channel of CNN feature maps contain the style information of an image, and hence they are used to translate feature maps of a source image into a normalized version based on a reference image for each channel. In addition to being able to match feature statistics, the ability to maintain spatial structures in feature maps with channel-wise normalization makes it appealing for tasks like segmentation.

Motivated by these observations, we propose to reduce domain differences at both low-level and high-level through channel-wise alignment. In particular, we normalize features of images from the source domain with those of images from the target domain by matching their channel-wise feature statistics. Nevertheless, such alignment is on a per image basis with each target sample serving as a reference for calibration. When multiple images exist in the target domain, a straightforward way is to enumerate all of them to cover all possible variations, which is computationally expensive. In contrast, we stochastically sample from the target domain for alignment. The randomization strategy is not only efficient, but more importantly, provides a form of regularization for training in similar spirit to stochastic depth [18], data transformation [19,20], and dropout [21].

To this end, we present, Dual Channel-wise Alignment Networks (DCAN), a simple yet effective framework optimized in an end-to-end manner. The main idea is leveraging images from the target domain for channel-wise alignment, which not only enables minimizing the low-level domain discrepancies in pixel space (*e.g.*, color, texture, lighting conditions, *etc.*), but also, simultaneously normalizes high-level feature maps of source images specific to those of target images for improved segmentation. Figure 1 gives an overview of the framework. In particular, we utilize an image generator to map an image from the source domain to multiple representations with the same content as the input but in different styles, determined by unlabeled images randomly selected from the target set. These synthesized images, resembling samples from the target domain, together with sampled target images, are further input into a segmentation network, in which channel-wise feature alignment is performed once more to refine features for the final segmentation task.

The key contributions of DCAN are summarized as follows: (1) we present an end-to-end learning framework, guided by feature statistics of images from the target domain, to synthesize new images as well as normalize features on-the-fly for unsupervised scene adaptation; (2) we demonstrate that channel-wise feature alignment, preserving spatial structures and semantic concepts, is a simple yet effective way to reduce domain shift in high-level feature maps. With this, our method departs from much recent and concurrent work, which uses adversarial training for distribution alignment; (3) we conduct extensive experiments by transferring models trained on synthetic segmentation benchmarks, *i.e.*, SYNTHIA [22] and GTA5 [23], to real urban scenes, CITYSCAPES [24], and demonstrate DCAN outperforms state-of-the-art methods with clear margins and it is compatible with several modern segmentation networks.

## 2 Related Work

There is a large body of work on domain adaptation (see [25,26] for a survey), and here we focus only on the most relevant literatures.

**Unsupervised Domain Adaptation**. Most existing work focuses on classification problems and falls into two categories: feature-level and pixel-level adaptation. Feature-level adaptation seeks to align features by either explicitly minimizing the distance measured by Maximum Mean Discrepancies (MMD) [7,27], covariances [6], *etc.*, between source and target distributions or implicitly optimizing adversarial loss functions in the forms of reversed gradient [28,29], domain confusion [30], or Generative Adversarial Network [8,9], such that features are domain-invariant. In contrast, pixel-level domain adaptation attempts to remove low-level differences like color and texture by stylizing source images to resemble target images [10,31,32,33]. Compared to a large amount of work on classification problems, limited effort has been made for semantic segmentation. In [9], adversarial training is utilized to align features in fully convolutional networks for segmentation, and the idea is further extended for both pixel-level and feature-level adaptation jointly using cycle consistency [11]. A curriculum learning strategy is proposed in [13] by leveraging information from global label distributions and local super-pixel distributions. Our work differs from previous work in two aspects: (1) we introduce channel-wise alignment for unsupervised scene adaption, which preserves spatial information and semantic information of each channel when normalizing high-level feature maps for alignment; (2) we avoid adversarial training, which "remains remarkably difficult to train" [34], yet achieves better performance.

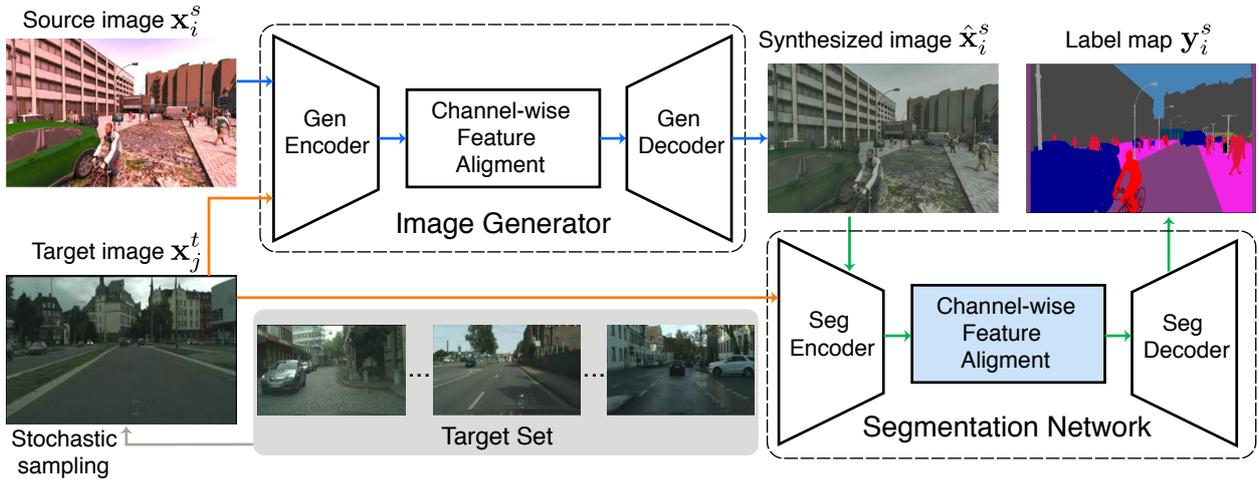

Fig. 1: An overview of the proposed framework. It contains an image generator and a segmentation network, in both of which channel-wise alignment is performed. The generator synthesizes a new image, reducing low-level appearance differences, which is further input to the semantic segmentation network. Features directly used for segmentation are refined before producing prediction maps. During testing, we turn off the alignment (shaped in blue) and the segmentation network can be readily applied.

**Image Synthesis**. Generative Adversarial Networks (GANs) [35], consisting of a generator and a discriminator optimized to compete with each other, are one of the most popular deep generative models for image synthesis [36,37,33]. Various prior information, including labels [38], text [39], attributes [40], images [41,42] has been explored to condition the generation process. GANs have also been further extended to the problem of image-to-image translation, which maps a given image to another one in a different style, using cycle consistency [43] or a shared latent space [44]. This line of work aims to learn a joint distribution of images from two domains using images from the marginal distributions of each domain. As previously mentioned, adversarial loss functions are hard to train, and hence generating high resolution images is still a challenging problem that could take days [45]. A different direction of image-to-image translation is neural style transfer [46,17,47,48,49]. Though style transfer can be seen as a special domain adaptation problem with each style as a domain [50], our goal in this work is different: we focus on unsupervised scene adaption, by jointly synthesizing images and performing segmentation with the help of images from the target domain for channel-wise distribution alignment.

## 3    Approach

Given labeled images from a source domain and unlabeled samples from a target domain, our goal is to reduce domain discrepancies at both pixel-level and feature-level. In particular, we leverage unlabeled target images for channel-wise alignment—synthesizing photo-realistic samples to appear as if from the target set, and simultaneously normalizing feature maps of source images, upon which segmentation classifiers directly rely. The resulting segmentation model can then be readily applied to the novel target domain. To this end, we consider each image from the target domain as a unique reference sample, whose feature representations are used to normalize those of images from the source domain. In addition, given an image from the source domain, instead of considering every single target image, we sample from the target set for alignment stochastically, serving as regularization to improve generalization. Figure 1 gives an overview of this framework.

More formally, let $\mathbf{X}^s = \{\mathbf{x}_i^s, \mathbf{y}_i^s\}_{i \in [N^s]}$ denote the source domain with $N^s$ images $\mathbf{x}_i^s \in \mathbb{R}^{3 \times H \times W}$ and the corresponding label maps $\mathbf{y}_i^s \in \{0,1\}^{C \times H \times W}$, where $H$ and $W$ represent the height and width of the image, respectively and $C$ denotes the number of classes. The target domain, on the other hand, has $N^t$ images $\mathbf{X}^t = \{\mathbf{x}_j^t\}_{j \in [N^t]}$ of the same resolution without labels. For each image $\mathbf{x}_i^s$ in the source domain, we randomly select one sample $\mathbf{x}_j^t$ from the target domain (we use one image here for the ease of description, but it can be a set of images as will be shown in experiments). A synthesized image $\hat{\mathbf{x}}_i^s$ is generated with the content of $\mathbf{x}_i^s$ and style of $\mathbf{x}_j^t$ by channel-wise alignment of feature statistics. This image is then fed into a segmentation network, where domain shift in high-level feature maps is further minimized for segmentation.

In the following, we first revisit channel-wise alignment (Sec 3.1), and then we present DCAN (Sec 3.2), which contains an image generator, synthesizing new images to minimize low-level differences like color and texture, and a segmentation network, refining high-level feature maps that are critical in the final segmentation task. Finally, we introduce the learning strategy (Sec. 3.3).

### 3.1 Channel-wise Feature Alignment

The mean and standard deviation of each channel in CNN feature maps have been shown to capture the style information of an image [16,15,17], and hence channel-wise alignment of feature maps is adopted for fast style transfer with a simple instance normalization step. Here, due to its effectiveness and simplicity, we use adaptive instance normalization [17], to match the mean and standard deviation of images from two different domains. In particular, given feature maps $F_i^s$ and $F_j^t$ of the same size $\mathbb{R}^{\hat{C} \times \hat{H} \times \hat{W}}$ ($\hat{C}$, $\hat{H}$, $\hat{W}$ represents the channel, height and width respectively) from the source and target domain, adaptive instance normalization $h$ produces a new representation of the source image as:

$$\hat{F}_i^s = h(F_i^s, F_j^t) = \sigma(F_j^t) \left( \frac{F_i^s - \mu(F_i^s)}{\sigma(F_i^s)} \right) + \mu(F_j^t), \tag{1}$$

$$\mu_c(F) = \frac{1}{\hat{H}\hat{W}} \sum_{h=1}^{\hat{H}} \sum_{w=1}^{\hat{W}} F_{chw}, \quad \sigma_c^2(F) = \frac{1}{\hat{H}\hat{W}} \sum_{h=1}^{\hat{H}} \sum_{w=1}^{\hat{W}} (F_{chw} - \mu_c(F))^2,$$

where $\mu_c$ and $\sigma_c$ denotes mean and variance across spatial dimensions for the $c$-th channel. This simple operation normalizes features of a source image to have similar statistics with those of a target image for each channel, which is appealing for segmentation tasks, since it is spatially invariant, *i.e.*, relative locations of pixels are fixed. In addition, such channel-wise alignment ensures semantic information like attributes encoded in different channels [14] is processed independently. In our work, we adopt channel-wise feature alignment in both our image generator for synthesizing photo-realistic samples, and segmentation network to refine features used for segmentation. Note that channel-wise feature alignment is generic and can be plugged into different layers of networks.

### 3.2  Dual Channel-wise Alignment Networks

**Image generator**. Our image generator contains an encoder and a decoder with channel-wise alignment in between. More specifically, the encoder, denoted as $f_{gen}$, is truncated from a pre-trained VGG19 network [51] by taking layers up till `relu4`. We fix the weights of the encoder, following [52,17], to map images $\mathbf{x}_i^s$ and $\mathbf{x}_j^t$ into fixed representations: $F_i^s = f_{gen}(\mathbf{x}_i^s)$ and $F_j^t = f_{gen}(\mathbf{x}_j^t)$, respectively. $F_i^s$ is further normalized to produce a new representation $\hat{F}_i^s$ according to Eqn. (1). Given the aligned source representation, a decoder, represented by $g_{gen}$, is applied to synthesize a new image $\hat{\mathbf{x}}_i^s = g_{gen}(\hat{F}_i^s)$, in the style of samples from the target set. This is achieved by minimizing the following image generation loss function:

$$\ell_{gen} = \|f_{gen}(\hat{\mathbf{x}}_i^s) - \hat{F}_i^s\|_2 + \sum_{l=1}^{4} \|G(f_{gen}^l(\hat{\mathbf{x}}_i^s)) - G(f_{gen}^l(\mathbf{x}_j^t))\|_2. \quad (2)$$

Here, the first term is the content loss measuring the discrepancies between features from the stylized image $\hat{\mathbf{x}}_i^s$ and the aligned features of the source image (weights of $f_{gen}$ are fixed), forcing the synthesized image to contain the same contents as the original one. The second term matches the style information, by penalizing the differences of Gram matrices between $\hat{\mathbf{x}}_i^s$ and the target image $\mathbf{x}_j^t$ using features from the first four layers (with $l$ denoting the layer index) in the encoder [46]. More specifically, given a reshaped feature map $F$ with its original channel, height and width being $\hat{C}$, $\hat{H}$, $\hat{W}$ respectively, the gram matrix can be computed as:

$$G(F) = \sum_{k=1}^{\hat{H}\hat{W}} F_{ik} F_{jk} \in \mathbb{R}^{\hat{C} \times \hat{C}}, \ F \in \mathbb{R}^{\hat{C} \times \hat{H}\hat{W}}.$$

**Segmentation network**. A new image $\hat{\mathbf{x}}_i^s$ synthesized with our generator, resembling target samples with similar low-level details like color, texture, lighting, *etc.*, is ready for semantic segmentation. Instead of sending $\hat{\mathbf{x}}_i^s$ to any off-the-shelf segmentation engine for the task, we leverage the target style image $\mathbf{x}_j^t$ once more to calibrate features of $\hat{\mathbf{x}}_i^s$ with channel-wise alignment, such that they possess similar statistics and its spatial information is preserved for segmentation. Here, the intuition is to remove undesired mismatches in higher-level feature maps that might still exist after minimizing low-level differences in the first stage. Therefore, DCAN explicitly performs another round of alignment in the segmentation network, refining features tailored for pixel-level segmentation. To this end, we divide a fully convolutional network (FCN) based model into an encoder $f_{seg}$ and a decoder $g_{seg}$, with alignment in between. In particular, the segmentation decoder produces a prediction map: $\mathbf{p}_i^s = g_{seg}(h(f_{seg}(\hat{\mathbf{x}}_i^s), f_{seg}(\mathbf{x}_j^t)))$ and the segmentation loss $\ell_{seg}$ takes the form:

$$\ell_{seg} = -\sum_{m=1}^{H \times W} \sum_{c=1}^{C} \mathbf{y}_i^{mc} \log(\mathbf{p}_i^{mc}), \quad (3)$$

which is essentially a multi-class cross-entropy loss summed over all pixels (superscript $s$ denoting the source domain is omitted here). Note that state-of-the-art segmentation networks like DeepLab [53], FCN [54], PSPNet [55], GCN[56], *etc.*, are usually built upon top-performing models on ImageNet like VGG [57] or ResNet [58]; these networks differ in depth but have similar configurations, *i.e.*, five groups of convolution. In this case, we utilize the first three convolution groups from a segmentation model as our encoder and the remaining part as the decoder. For encoder-decoder based segmentation networks like SegNet [59], the simple idea could be directly applied.

In summary, DCAN works in the following way: given a source image, a target image is randomly selected whose style information is used for dual channel-wise alignment in both image synthesis and segmentation phases. The image generator first synthesizes a new image on-the-fly to appear similar as samples from the target domain, reducing low-level domain discrepancies in pixel space (*e.g.*, color, texture, lighting conditions, *etc.*), which is further input into the segmentation network. In the segmentation model, features from the synthesized image are further normalized specific to the sampled target image while preserving spatial structures and semantic information before producing label maps.

At test time, a novel image from the target domain is input into the segmentation network (segmentation encoder and then decoder) to predict its semantic map. The channel-wise feature alignment in the segmentation network is turned off since the network is already trained to match the feature statistics between two domains and thus can be directly applied for testing as shown in Figure 1.

### 3.3 Optimization

One could train the framework by selecting each sample in the source domain and normalizing it with the style information of each image in the target domain, which leads to $N^t$ copies of the original image; the new dataset $\hat{\mathbf{X}}^s$ with the size of $N^s N^t$ can then be used for training by minimizing:

$$\mathcal{L} = \frac{1}{N^s} \sum_{i=1}^{N^s} \frac{1}{N^t} \sum_{j=1}^{N^t} (\ell_{seg}(\mathbf{x}_i^s, \mathbf{x}_j^t, \mathbf{y}_i^s; \Theta_{seg}) + \lambda \ell_{gen}(\mathbf{x}_i^s, \mathbf{x}_j^t; \Theta_{gen})), \quad (4)$$

where $\Theta_{seg}$ and $\Theta_{gen}$ denote the parameters for the segmentation network and the image generator, respectively, and $\lambda$ balances the two losses. However, enumerating all targets would be computationally expensive, as the cost grows linearly with the number of images in the target domain. It is worth noting that when there are infinite target images, Eqn (4) can be re-written as:

$$\mathcal{L} = \frac{1}{N^s} \sum_{i=1}^{N^s} \mathbb{E}_{\mathbf{x}_j^t \sim \mathbf{X}^t} [\ell_{seg}(\mathbf{x}_i^s, \mathbf{x}_j^t, \mathbf{y}_i^s; \Theta_{seg}) + \lambda \ell_{gen}(\mathbf{x}_i^s, \mathbf{x}_j^t; \Theta_{gen})]. \qquad (5)$$

Here, the expected mean can be computed by stochastic sampling during training. The intuition is to introduce "uncertainties" to the learning processes as opposed to summing over all target styles deterministically, making the derived model more robust to noise and to generalize better on the target domain. It is a type of regularization similar in spirit to SGD for fast convergence [60], stochastic depth [18] and dropout [21,61,62]. Another way to view this is randomized data augmentation to improve generalization ability [63,51]. Unlike PixelDA [10] which generates new samples conditioned on a noise vector, we augment data using feature statistics of images randomly sampled from the target domain. It is also worth noting that the idea of sampling is in line with stochastic gradient descent, which loops over the training set by sampling batches of images, and hence can be easily implemented in current deep learning frameworks.

## 4  Experiments

In this section, we first introduce the experimental setup and implementation details. Then, extensive experimental results are presented to demonstrate the effectiveness of our method. Finally, an ablation study is conducted to evaluate the contribution of different components of DCAN.

### 4.1  Experimental Setup

**Datasets and evaluation metrics**. We train DCAN on two source datasets, SYNTHIA [22] and GTA5 [23] respectively, and then evaluate the models on CITYSCAPES [24]. CITYSCAPES is a real-world dataset, capturing street scenes of 50 different cities, totaling 5,000 images with pixel-level labels. The dataset is divided into a training set with 2,975 images, a validation set with 500 images and a testing set with 1,525 images. SYNTHIA is a large-scale synthetic dataset automatically generated for semantic segmentation of urban scenes. As in [13,9], we utilize SYNTHIA-RAND-CITYSCAPES, a subset that contains 9,400 images paired with CITYSCAPES, sharing 16 common classes. We randomly select 100 images for validation and use the remaining 9,300 images for training. GTA5 contains 24,966 high-resolution images, automatically annotated into 19 classes. The dataset is rendered from a modern computer game, Grand Theft Auto V, with labels fully compatible with those of CITYSCAPES. We randomly pick 1,000 images for validation and use the remaining 23,966 images for training.

Following [13,9], to train our model, we utilize *labeled* images from the training set of either SYNTHIA or GTA5, as well as *unlabeled* images from the training set of CITYSCAPES serving as references for distribution alignment. Then we evaluate the segmentation model on the validation set of CITYSCAPES, and report mean intersection-over-union (mIoU) to measure the performance. These two adaptation settings are denoted as SYNTHIA → CITYSCAPES and GTA5 → CITYSCAPES, respectively.

**Network architectures**. For the image generator, its encoder is based on a VGG19 network; the detailed architecture of the decoder can be found in the supplemental material. To verify the effectiveness of DCAN in state-of-the-art segmentation networks, we experiment with three top-performing architectures, FCN-8s-VGG16 [54], FCN-8s-ResNet101, and PSPNet [55]. In particular, FCN8s-VGG16 and FCN8s-ResNet101 respectively adapt a pre-trained VGG16 and a ResNet101 network into fully convolutional networks and use skip connections for detailed segmentations. PSPNet is built upon a ResNet50 model with a novel pyramid pooling module to obtain representations of multiple sub-regions for per-pixel prediction [55]. These networks are pre-trained on ImageNet.

**Implementation details**. We adopt PyTorch for implementation and use SGD as the optimizer with a momentum of 0.99. The learning rate is fixed to $1e-3$ for both FCN8s-ResNet101 and PSPNet, and $1e-5$ for FCN8s-VGG16. We adopt a batch size of three and optimize for $100,000$ iterations, and we fix $\lambda$ to 0.1. Given each sample in the training set, we randomly sample 2 images and 1 image from the target image set for experiments on SYNTHIA and GTA5 respectively. This is to achieve efficient training on GTA5, for its size is three times larger than SYNTHIA, and we will analyze the effect of the number of sampled images below. We use a crop of $512 \times 1024$ during training, and for evaluation we upsample the prediction map by a factor of 2 and then evaluate mIoU.

### 4.2 Main Results

We compare DCAN to state-of-the-art methods on unsupervised domain adaptation for semantic segmentation, including "FCN in the wild" [9] and "Curriculum Adaptation" [13]. In particular, FCN in the wild uses an adversarial loss to align fully connected layers (adapted to convolution layers) of a VGG16 model, and additionally leverages multiple instance learning to transfer spatial layout [9]. Curriculum Adaptation infers properties of the target domain using label distributions of images and superpixels [13]. The results of SYNTHIA → CITYSCAPES and GTA5 → CITYSCAPES are summarized in Table 1.

We observe that these domain adaptation methods, although different in design, can indeed lead to improvements over the source only method (denoted as source), which simply trains a model on the source domain and then directly applies it to the target domain. In particular, DCAN outperforms its corresponding source only baseline with clear margins, around 8 and 9 absolute percentage points, using all three different networks on both datasets. This confirms the effectiveness of DCAN, which not only reduces domain differences for

Table 1: Results and comparisons on CITYSCAPES when adapted from SYNTHIA and GTA5, respectively. Here, "Source" denotes source only methods, "Oracle" denotes results from supervised training, and A, B, C represent FCN8s-VGG16, FCN8s-ResNet101 and PSPNet. A/d uses dilation in VGG16 for segmentation.

SYNTHIA → CITYSCAPES

| Method | network | road | sidewalk | building | wall | fence | pole | traffic light | traffic sign | vegetation | sky | person | rider | car | bus | motorbike | bike | mIOU | mIOU gain |
|---|---|---|---|---|---|---|---|---|---|---|---|---|---|---|---|---|---|---|---|
| Source [9] | A/d | 6.40 | 17.7 | 29.7 | 1.20 | 0.00 | 15.1 | 0.00 | 7.20 | 30.3 | 66.8 | 51.1 | 1.50 | 47.3 | 3.90 | 0.10 | 0.00 | 17.4 | - |
| [9] | A/d | 11.5 | 19.6 | 30.8 | 4.40 | 0.00 | 20.3 | 0.10 | 11.7 | 42.3 | 68.7 | 51.2 | 3.80 | 54.0 | 3.20 | 0.20 | 0.60 | 20.2 | 2.80 |
| Source [13] | A | 5.60 | 11.2 | 59.6 | 8.00 | 0.50 | 21.5 | 8.00 | 5.30 | 72.4 | 75.6 | 35.1 | 9.00 | 23.6 | 4.50 | 0.50 | 18.0 | 22.0 | - |
| [13] | A | 65.2 | 26.1 | 74.9 | 0.10 | 0.50 | 10.7 | 3.50 | 3.00 | 76.1 | 70.6 | 47.1 | 8.20 | 43.2 | 20.7 | 0.70 | 13.1 | 29.0 | 7.00 |
| Source | A | 10.8 | 11.4 | 66.6 | 1.60 | 0.10 | 16.9 | 5.50 | 14.1 | 74.2 | 76.2 | 46.0 | 11.5 | 45.4 | 15.1 | 6.00 | 13.4 | 25.9 | - |
| DCAN | A | 79.9 | 30.4 | 70.8 | 1.60 | 0.60 | 22.3 | 6.70 | 23.0 | 76.9 | 73.9 | 41.9 | 16.7 | 61.7 | 11.5 | 10.3 | 38.6 | **35.4** | **9.5** |
| Source | B | 57.9 | 17.0 | 72.7 | 0.20 | 0.00 | 10.4 | 0.00 | 0.00 | 73.5 | 75.4 | 37.8 | 9.30 | 59.3 | 21.7 | 0.40 | 12.3 | 28.0 | - |
| DCAN | B | 81.5 | 33.4 | 72.4 | 7.90 | 0.20 | 20.0 | 8.60 | 10.5 | 71.0 | 68.7 | 51.5 | 18.7 | 75.3 | 22.7 | 12.8 | 28.1 | **36.5** | **8.5** |
| Source | C | 56.0 | 24.6 | 76.5 | 5.00 | 0.20 | 19.0 | 5.70 | 7.80 | 77.5 | 78.9 | 44.7 | 7.70 | 35.3 | 7.90 | 1.50 | 24.0 | 29.5 | - |
| DCAN | C | 82.8 | 36.4 | 75.7 | 5.08 | 0.06 | 25.8 | 8.04 | 18.7 | 74.7 | 76.9 | 51.1 | 15.9 | 77.7 | 24.8 | 4.11 | 37.3 | **38.4** | **8.9** |
| Oracle | A | 96.4 | 70.3 | 85.9 | 44.4 | 35.8 | 31.5 | 41.5 | 54.2 | 87.5 | 88.9 | 64.1 | 40.8 | 88.5 | 66.1 | 35.5 | 60.3 | 62.0 | - |
|  | B | 97.3 | 76.7 | 88.1 | 44.4 | 46.9 | 35.3 | 44.5 | 55.9 | 88.6 | 91.2 | 67.7 | 41.6 | 89.9 | 73.3 | 44.7 | 63.1 | 65.6 | - |
|  | C | 97.8 | 78.6 | 89.6 | 56.7 | 57.8 | 39.9 | 61.3 | 65.2 | 89.9 | 91.5 | 73.4 | 56.0 | 89.9 | 84.1 | 54.2 | 69.5 | 72.2 | - |

GTA5 → CITYSCAPES

| Method | network | road | sidewalk | building | wall | fence | pole | traffic light | traffic sign | vegetation | terrain | sky | person | rider | car | truck | bus | train | motorbike | bike | mIOU | mIOU gain |
|---|---|---|---|---|---|---|---|---|---|---|---|---|---|---|---|---|---|---|---|---|---|---|
| Source [9] | A/d | 31.9 | 18.9 | 47.7 | 7.40 | 3.10 | 16.0 | 10.4 | 1.00 | 76.5 | 13.0 | 58.9 | 36.0 | 1.00 | 67.1 | 9.50 | 3.70 | 0.00 | 0.00 | 0.00 | 21.2 | - |
| [9] | A/d | 70.4 | 32.4 | 62.1 | 14.9 | 5.40 | 10.9 | 14.2 | 2.70 | 79.2 | 21.3 | 64.6 | 44.1 | 4.20 | 70.4 | 8.00 | 7.30 | 0.00 | 3.50 | 0.00 | 27.1 | 5.90 |
| Source [13] | A | 18.1 | 6.80 | 64.1 | 7.30 | 8.70 | 21.0 | 14.9 | 16.8 | 45.9 | 2.40 | 64.4 | 41.6 | 17.5 | 55.3 | 8.40 | 5.0 | 6.90 | 4.30 | 13.8 | 22.3 | - |
| [13] | A | 74.9 | 22.0 | 71.7 | 6.00 | 11.9 | 8.40 | 16.3 | 11.1 | 75.7 | 13.3 | 66.5 | 38.0 | 9.30 | 55.2 | 18.8 | 18.9 | 0.00 | 16.8 | 16.6 | 28.9 | 6.6 |
| Source | A | 72.5 | 25.1 | 71.2 | 6.60 | 13.4 | 12.3 | 11.0 | 4.70 | 76.1 | 16.4 | 67.7 | 43.1 | 8.00 | 70.4 | 11.3 | 4.80 | 0.00 | 13.9 | 0.40 | 27.8 | - |
| DCAN | A | 82.3 | 26.7 | 77.4 | 23.7 | 20.5 | 20.4 | 30.3 | 15.9 | 80.9 | 25.4 | 69.5 | 52.6 | 11.1 | 79.6 | 24.9 | 21.2 | 1.30 | 17.0 | 6.70 | **36.2** | **8.4** |
| Source | B | 44.5 | 12.7 | 71.1 | 9.40 | 17.7 | 15.3 | 24.3 | 11.9 | 80.5 | 14.3 | 80.0 | 50.3 | 7.70 | 45.4 | 30.5 | 30.8 | 5.50 | 9.80 | 3.50 | 29.8 | - |
| DCAN | B | 88.5 | 37.4 | 79.3 | 24.8 | 16.5 | 21.3 | 26.3 | 17.4 | 80.8 | 30.9 | 77.6 | 50.2 | 19.2 | 77.7 | 21.6 | 27.1 | 2.70 | 14.3 | 18.1 | **38.5** | **8.7** |
| Source | C | 69.9 | 22.3 | 75.6 | 15.8 | 20.1 | 18.8 | 28.2 | 17.1 | 75.6 | 8.00 | 73.5 | 55.0 | 2.90 | 66.9 | 34.4 | 30.8 | 0.00 | 18.4 | 0.00 | 33.3 | - |
| DCAN | C | 85.0 | 30.8 | 81.3 | 25.8 | 21.2 | 22.2 | 25.4 | 26.6 | 83.4 | 36.7 | 76.2 | 58.9 | 24.9 | 80.7 | 29.5 | 42.9 | 2.50 | 26.9 | 11.6 | **41.7** | **8.4** |
| Oracle | A | 96.4 | 70.3 | 85.9 | 44.4 | 35.8 | 31.5 | 41.5 | 54.2 | 87.5 | 51.9 | 88.9 | 64.1 | 40.8 | 88.5 | 55.8 | 66.1 | 44.9 | 35.5 | 60.3 | 60.2 | - |
|  | B | 97.3 | 76.7 | 88.1 | 44.4 | 46.9 | 35.3 | 44.5 | 55.9 | 88.6 | 55.9 | 91.2 | 67.7 | 41.6 | 89.9 | 60.1 | 73.3 | 54.4 | 44.7 | 63.1 | 64.2 | - |
|  | C | 97.8 | 78.6 | 89.6 | 56.7 | 57.8 | 39.9 | 61.3 | 65.2 | 89.9 | 58.9 | 91.5 | 73.4 | 56.0 | 89.9 | 75.8 | 84.1 | 78.8 | 54.2 | 69.5 | 72.0 | - |

improved performance but also is general for multiple network architectures. Furthermore, with PSPNet, DCAN achieves 41.7% and 38.4% on CITYSCAPES when adapted from GTA5 and SYNTHIA, respectively. Compared to [9,13], with the same backbone VGG16 architecture, DCAN offers the best mIoU value as well the largest relative mIoU gain (9.5% and 8.4% trained from SYNTHIA and GTA5 respectively). Note that although the backbone network is the same, source only baselines are different due to different experimental settings. A dilated VGG16 network is adopted in [9] and the network is additionally pre-trained on PASCAL-CONTEXT in [13]. In addition, it uses a crop size of 320 × 640 during training. Our model is initialized on ImageNet and we choose 512 × 1024 for training since large resolution offers better performance as observed in [55], which is also consistent with state-of-the-art supervised methods on CITYSCAPES [24]. It is worth noting that DCAN improves a stronger baseline by 36% relatively (25.9% to 35.4%). With the same image size as in [13], DCAN improves the source only baseline from 23.6% to 33.0% (*v.s.*, 22.0% to 29.0% in [13]; see Table 2).

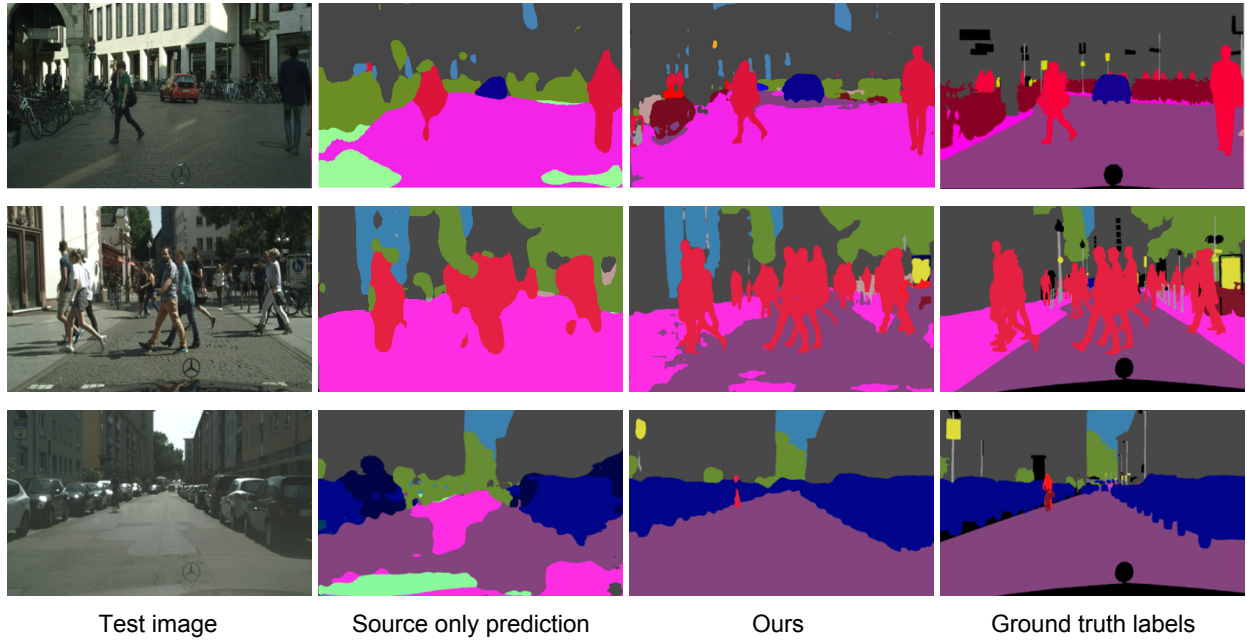

Fig. 2: Sampled prediction results of PSPNet and its corresponding source only model under the GTA5 → CITYSCAPES setting using testing images from CITYSCAPES. Our model effectively improves the generalization ability of the trained segmentation network.

Among three different networks, PSPNet gives the best results on both datasets, mainly resulting from the pyramid pooling module that considers difference scales. Figure 2 illustrates sampled results of PSPNet under the GTA5 → CITYSCAPES setting, and its comparison with the source only method. Comparing across datasets, models trained on GTA5 produce better accuracies than those learned from SYNTHIA. The reasons are two-folds: (1) a large number of images from SYNTHIA are rendered at night, incurring significant domain differences since images from CITYSCAPES are captured during day time; (2) there are more training samples in GTA5. In addition, oracle results, which are produced with traditional supervised training using annotations from the target domain, are also listed for reference. We can see there is still significant performance gaps between domain adaptation methods and oracle supervised training, highlighting the challenging nature of this problem.

### 4.3  Discussions

In this section, we run a number of experiments to analyze DCAN in the SYNTHIA → CITYSCAPES setting, and provide corresponding results and discussions.

| Synthia → Cityscapes | | | |
|---|---|---|---|
| Resolution | Method | mIoU | gain |
| 256×512 | Source | 21.2 | |
| | DCAN | 29.6 | 8.4 |
| 320×640 | Source | 23.6 | |
| | DCAN | 33.0 | 9.4 |
| 512×1024 | Source | 25.9 | |
| | DCAN | 35.4 | 9.5 |

Table 2: Results of FCN8s-VGG16 using three different image resolutions.

| Synthia → Cityscapes | |
|---|---|
| Method | mIoU |
| CycleGAN [43] | 30.4 |
| CycleGAN w. FeatureAlignment | 31.7 |
| UNIT [52] | 31.6 |
| UNIT w. FeatureAlignment | 32.7 |
| DCAN w/o FeatureAlignment | 33.8 |
| DCAN (two stage) | 33.7 |
| DCAN (end-to-end) | 35.4 |

Table 3: Training with and without feature alignment in FCN8s-VGG16 using different image synthesis methods.

**Image resolution**. As previously mentioned, top performing approaches on Cityscapes typically use a high resolution for improved performance [24]. For example, GCN and FRRN utilize a resolution of 800×800 [56] and 512×1024 [64], respectively. Here, we report the results of DCAN adapted from Synthia using FCN8s-VGG16 with three different resolutions, and compare with the corresponding source only method in Table 2. DCAN offers significant performance gains for all resolutions, and a larger resolution is indeed better for unsupervised domain adaptation.

**Different image synthesis methods**. We compare with two different image synthesis methods: (1) CycleGAN [43] and (2) UNIT [44], both of which attempt to learn a distribution mapping function between two domains. Once the mapping function is learned, images from the source domain can be translated to the style of the target domain. Therefore, we use the translated images from the source domain to train the segmentation network. Table 3 presents the results. For fair comparisons, we compare them under two settings, with and without the channel-wise feature alignment in the segmentation network. DCAN achieves better results than both GAN-based image synthesis methods in both scenarios. To justify the advantage of an end-to-end framework, we also compare with a two-stage training strategy, which simply trains a segmentation network using pre-synthesized images without end-to-end training. In this case, image synthesis is not optimized using gradients from the segmentation network. DCAN improves the two-stage training by 1.7% mIOU, demonstrating the importance of guiding the synthesis process with useful information from the final task.

Figure 3 further compares images produced by different synthesis methods. DCAN is able to generate images that conform to the style of images from the target set, containing fewer artifacts than CycleGAN and UNIT. In addition, both CycleGAN and UNIT seek to align distributions at a dataset level, and once the mapping is learned, the translation from the source to the target is fixed (a fixed output given an input image). Learning such a transformation function on high resolution images is a non-trivial task and might not perfectly cover all possible variations. Instead, DCAN performs image translation at an instance level, and in the regime of stochastic sampling, it is able to cover sufficient styles from the target set for distribution alignment. It is also worth noting that feature alignment can improve segmentation results regardless of synthesis methods. We also experimented with other GAN-based approaches like PixelDA [10] for image synthesis; however, conditioning on a noise vector rather than label maps [65] fails to produce photo-realistic images in high resolution.

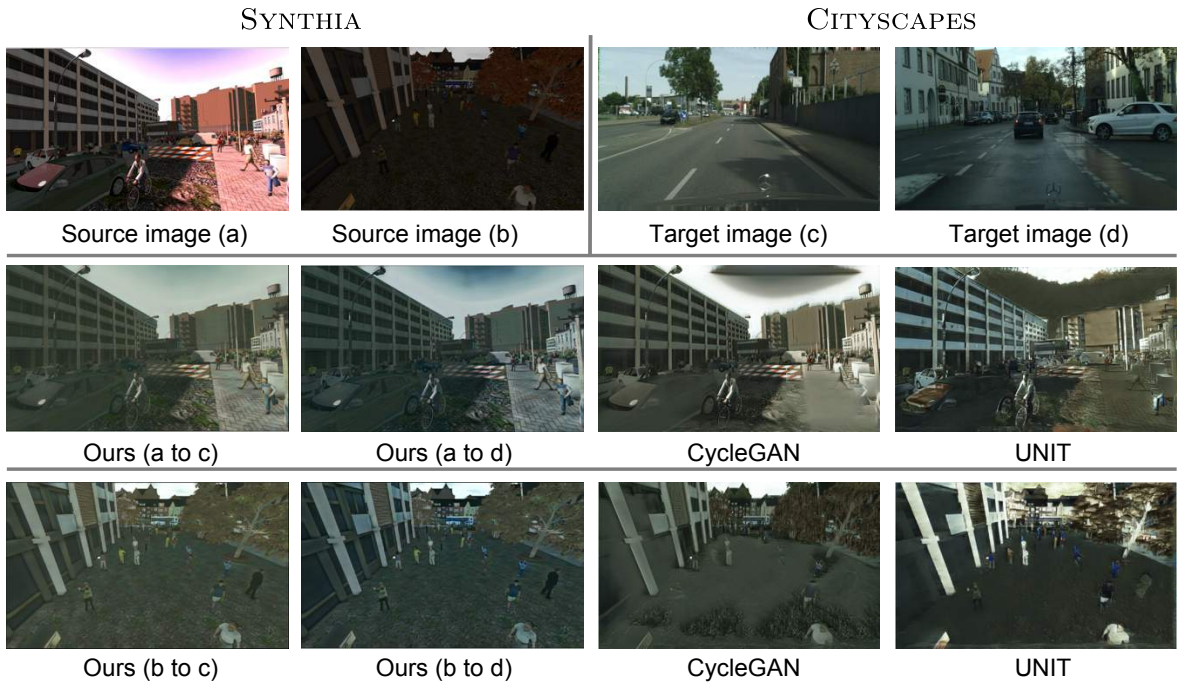

Fig. 3: Images from SYNTHIA synthesized in the style of CITYSCAPES with CycleGAN [43], UNIT [44] and DCAN.

| SYNTHIA → CITYSCAPES | |
|---|---|
| Alignment Method | mIoU |
| ADDA [8] | 34.0 |
| Ours-w/o alignment | 33.8 |
| Ours–`Conv2` | 34.0 |
| Ours–`Conv4` | 34.4 |
| Ours–`Conv6` | 33.2 |
| Ours–`Conv7` | 32.7 |
| Ours–`Conv3` | 35.4 |

Table 4: Comparisons of different feature alignment methods in the segmentation network.

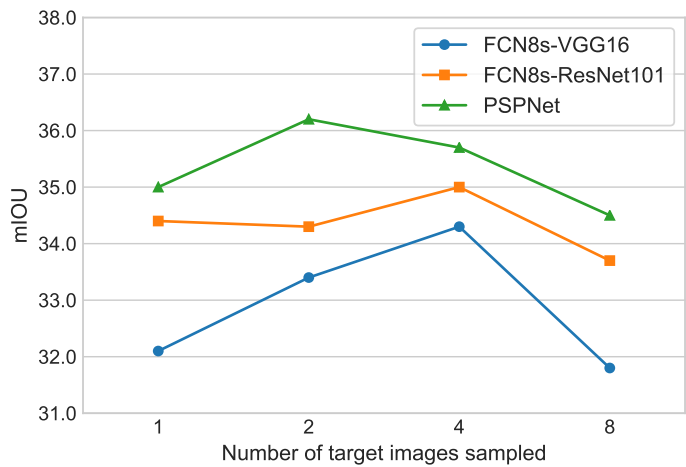

Fig. 4: Effect of using different number of target images for each training sample.

**Channel-wise feature alignment for segmentation**. We now analyze the effect of channel-wise alignment in the segmentation network (Table 4) with FCN8s-VGG16. We compare with Adversarial Discriminative Domain Adaptation [8], which leverages an adversarial loss to make features from two domains indistinguishable without considering spatial structures explicitly. DCAN outperforms ADDA by 1.4%, and also converges faster during training. We also implemented MMD [7] and CORAL [6] loss to align features, but their results are worse than source only methods. This is consistent with observations in [13]. We further investigate where to align in the segmentation network and found that alignment after the `Conv3` layer gives the best results, possibly due to it contains both sufficient number of channels and relatively large feature maps. In addition, aligning features maps with more detailed spatial information (`Conv2` and `Conv4`) is also better than `Conv6` and `Conv7` (convolution layers adapted from fully connected layers, whose feature maps are smaller). This confirms the importance to consider detailed spatial information explicitly for alignment.

**Number of target images sampled**. We also evaluate how the number of sampled target images affects the performance. Since enumerating around 3,000 samples for each image in the training set is computationally prohibitive, we create a pseudo-target set with 8 images randomly selected from 8 cities in CITYSCAPES. This is to ensure there are variations among the targets and it is computationally feasible for enumerating all targets. We then analyze the effect of the number of target images used during training by randomly selecting 1, 2, 4 samples from SYNTHIA. Figure 4 presents the results. We observe that stochastically selecting from the target set is better than using all of them for all three networks. This might result from two reasons: (1) translating one image to multiple different representations in one-shot is hard to optimize; (2) stochastic sampling acts as regularization to improve generalization, which is similar to the case that stochastic gradient is better than full batch gradient descent. Interestingly, for PSPNet and FCN8s-ResNet101, sampling one image achieves competitive results, and this is very appealing when the number of samples in the target domain is limited.

## 5 Conclusion

In this paper, we have presented, DCAN, a simple yet effective approach to reduce domain shift at both pixel-level and feature-level for unsupervised scene adaptation. In particular, our framework leverages channel-wise feature alignment in both the image generator for synthesizing photo-realistic samples, appearing as if drawn from the target set, and the segmentation network, which simultaneously normalizes feature maps of source images. In contrast to recent work that makes extensive use of adversarial training, our framework is lightweight and easy to train. We conducted extensive experiments by transferring models learned on synthetic segmentation datasets to real urban scenes, and demonstrated the effectiveness of DCAN over state-of-the-art methods and its compatibility with modern segmentation networks.